\documentclass{svproc}
\usepackage{subfigure}
\usepackage{graphicx}
\usepackage{url}

\begin{document}
\mainmatter              

\title{An Improved Algorithm of Robot Path Planning in Complex Environment Based on Double DQN}
\titlerunning{An Improved Algorithm Based On DDQN}  
%
\author{Fei Zhang\inst{1} \and Chaochen Gu*\inst{1} \and
Feng Yang\inst{2}}
\authorrunning{Fei Zhang et al.} 
%
\tocauthor{Fei Zhang, Chaochen Gu, and Feng Yang}
\institute{Shanghai Jiao Tong University, Shanghai, 200240, China,\\
\email{jacygu@sjtu.edu.cn}
\and
Northwestern Polytechnical University, ShaanXi, 710129 China}

\maketitle              

\begin{abstract}
Deep Q Network (DQN) has several limitations when applied in planning a path in environment with a number of dilemmas according to our experiment. The reward function may be hard to model, and successful experience transitions are difficult to find in experience replay. In this context, this paper proposes an improved Double DQN (DDQN) to solve the problem by reference to A* and Rapidly-Exploring Random Tree (RRT). In order to achieve the rich experiments in experience replay, the initialization of robot in each training round is redefined based on RRT strategy. In addition, reward for the free positions is specially designed to accelerate the learning process according to the definition of position cost in A*. The simulation experimental results validate the efficiency of the improved DDQN, and robot could successfully learn the ability of obstacle avoidance and optimal path planning in which DQN or DDQN has no effect.
\keywords{Path Planning, Double DQN, Obstacle Avoidance}
\end{abstract}
\section{Introduction}
As one of important technical issues of robotic research, path planning aims to drive robot to find an optimal and safe path from the initial state to the target state in an environment with some obstacles. Up to now, classical algorithms for path planning such as A* \cite{b1} and Rapidly-Exploring Random Tree (RRT) \cite{b3} have been successfully applied in abundant path planning projects. However, time-consuming computation of high dimensional map \cite{b4} and low generalization capacity have become major shortages for them.

Deep Q Network (DQN) is an evolutionary algorithm based on the combination of deep learning and reinforcement learning. With the excellent performance in computer games \cite{b7,b8}, DQN has been gradually applied to robotic path planning and made considerable success \cite{b14,b16,b17}. Additionally, advances in DQN, typically is Double DQN (DDQN)  \cite{b11}, have enabled agent to better learn policy through the complex environment.

Unlike the experiments made by \cite{b14,b16,b17}, we focus on the problem of planning a relatively optimal path in complex grid maps based on DDQN and the extent of the complexity is decided by the number of dilemmas in map. Three expected goals are set for the trained DDQN: i) Finding an optimal path from a given start to a given end; ii) Searching an optimal path from random start to the given end; iii) Obtaining an optimal path from the given start to the given end with few distractions in maps. Unfortunately, the trained model of DDQN in the designed maps was failed after a number of trials. Two major problems are found during the training process. One is that the robot could not learn new experience because of the same initial state after ending in each round, which happens when robot reaches the end or obstacle. The other is that identical reward of free space causes difficulty in convergence of the model.

In order to solve the two problems mentioned above, this paper proposes some improvements on DDQN inspired by the classical algorithms of path planning. The set of initial state was changed with a certain probability after each iteration by the reference of RRT. In addition, the reward to free space is specially designed based on A* to accelerate the model's convergence. During testing experiment, the robot is able to successfully achieve the expected goal i), ii) and iii) based on such improved approach. The testing results show that the robot is capable of learning obstacles avoidance and finding path in grid map.
\section{Mathematic Model in Robot Path Planning}
\subsection{Designed Planning Area Description}
 Some grid maps for path planning based on reinforcement learning \cite{b18,b19} as showed in Fig.1, are generated with random settings of the start point, the end point and the free points. However, such weakly-designed environment might cause several problems. Firstly, the dilemma for the robot is hard to appear on the maps since the obstacles are mostly scattered points \cite{b18}. Secondly, the distance between the start point and the end point might be so small that planning path on such map has little significance \cite{b19}. Thus, in our work the obstacles are specially designed to form some dilemmas for robot.

\begin{figure}[htbp]
  \centering
  \subfigure[]{
    \label{fig:subfig:a} 
    \includegraphics[width=1.2in]{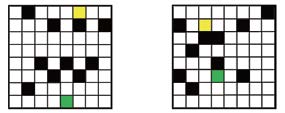}
  }
  \subfigure[]{
    \label{fig:subfig:b} 
    \includegraphics[width=1.2in]{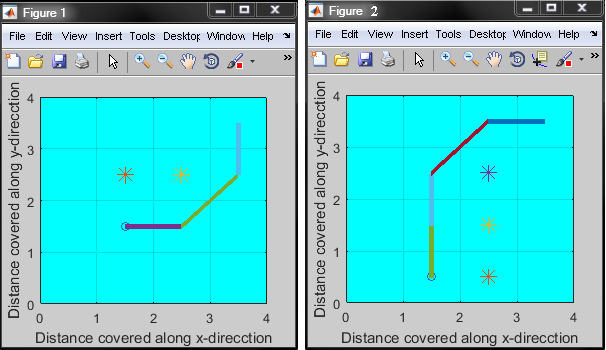}
  }
  \caption{A display of some weakly-designed grid maps for robotic motivation.}
\end{figure}

In this paper, the working space of robot can be represented by two-dimensional grid figure with equal size. Four different values are endowed to grids so as to distinguish the start position, the end position, the obstacles and the free space. As the experiment map in Fig. 2, the working space is divided into 20$\times$20 parts with the grid size. The green gird refers to the start position of robot and the blue one refers to the end position. The black grids represent the obstacles and the white ones are free positions. Cartesian coordinate system is set for the working space to represent the index or position of grids, so the position of a gird might be represented by the coordinate of the grid.
\begin{figure}[htbp]
\centerline{\includegraphics[width=0.45\textwidth]{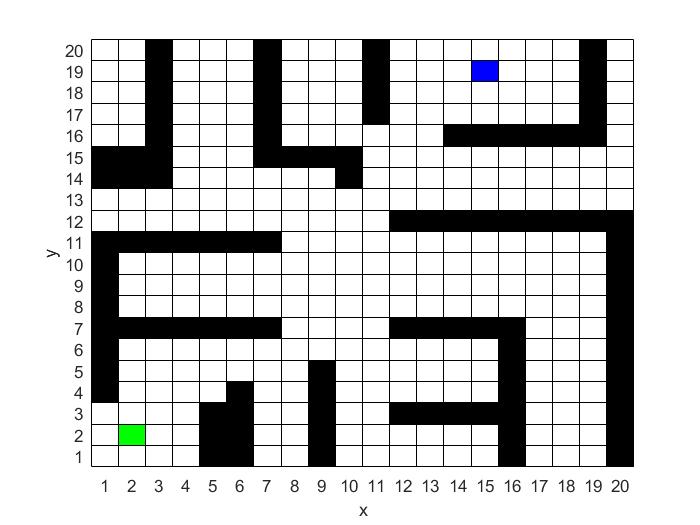}}
\caption{Two-dimensional working space of robot in our work.}
\label{Fig. 1}
\end{figure}

\subsection{Action Space Description}
The work of robot is to plan an optimal path without reaching any obstacle from the given start position to the given end position. In this paper, robot is simply allowed to move one step from its current position to the 8-adjacent grids of its current grid(see in Fig. 3). 8 actions are noted as value from 1 to 8. Thus a set of actions can defined as $Actions=\{1,2,3,4,5,6,7,8\}$.
\section{Path Planning Based On DDQN}
\subsection{Framework of DDQN}
The idea of DQN algorithm is to use the outputs of a multi-layer neural network for a given state $s$ and action $a$ to represent the action values $Q(s,a;\theta)$, where $\theta$ is the parameters of the network. To improve the performance of DQN, experience relay and target network are used in the training process of DQN \cite{b7}. In experience relay, observed transitions are stored and randomly sampled when updating the network. For the target network with parameters $\theta^-$, the construction of the target network is the same as the online network. $\theta^-$ simply copy from $\theta$ every specific $N$ steps and remain fixed during the other time. In order to reduce overoptimistic value estimates \cite{b11}, Double DQN redefines the target to train the online network as

\begin{eqnarray}
Y_t=R_{t+1}+{\gamma}Q(s_{t+1},\mathop{argmax}\limits_{a}Q(s_{t+1},a;\theta_t))\label{eq}
\end{eqnarray}

where $R$ refers to the reward that agent could obtain after choosing action $a$ in state $s$. Furthermore, the corresponding loss function to train the online defines as

\begin{eqnarray}
L_t(\theta_t)=E[(Y_t-Q(s_t,a;\theta_t))^2]\label{eq}
\end{eqnarray}

The overall framework of DDQN for robot path planning is showed in Fig. 3, which present an interacting process between agent (robot) and environment (map) by using a specially-designed deep Q network. The inputs of the neural network are the observed transitions $(s_t,a_t,r_t,s_{t+1})$ randomly mini-batched from experience relay, where $s_t$ is the robot's current position, $a_t$ is the action number that robot made from $Actions$, $r_t$ is the reward that robot obtained from the environment and $s_{t+1}$ is the next position of robot after executing action $a_t$. The outputs of neural networks are the 8 action values $Q(s,a;\theta)$ corresponding to each possible action from $Actions$. During the training process, robot selects the following actions by using $\varepsilon$-greedy strategy to make balance between exploration and exploitation \cite{b8}.
\begin{figure}[htb]
\centerline{\includegraphics[width=0.5\textwidth]{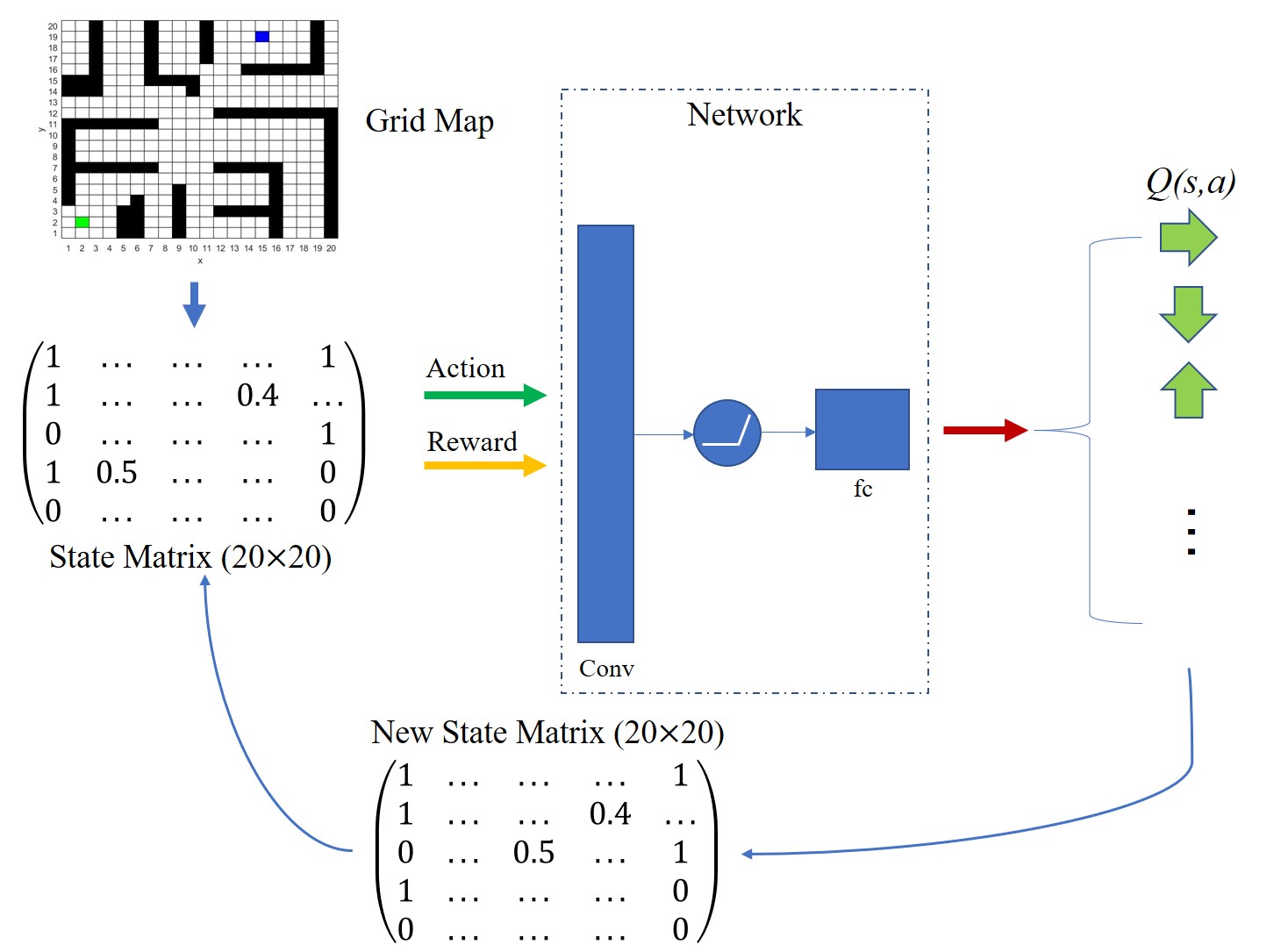}}
\caption{The framework of DDQN on path planning.}\label{Fig. 3}
\end{figure}
\subsection{Network Architectures}
In this work, the essence of grid map could be defined as a Matrix with special values. Thus the mini-batched matrixes of image with size of 20$\times$20 can be directly deemed as the inputs of the network. The whole construction of network showed in Fig. 3 mainly consists of one convolutional layer and one fully-connected layer. For the convolutional layer, the size of filter is 3$\times$3 and scan the inputs with stride=2. For the fully-connected layer, the outputs are 8 vectors corresponding to each action. One hidden layer is followed by RELU function. The network is trained with a Adams optimizer of learning rate 0.01.
\section{New Learning Strategy based on DDQN}
\subsection{Change in Initialization }
 In this paper, an ending in each round is defined as the time when robot chooses obstacles to move forward or reaches the end position as in \cite{b21,b22,b23}. The initialization of robot's position occurs after each ending. Generally, robot's position is supposed to be set to the given start position as in \cite{b21,b22,b23}. However, the training process based on DDQN failed every time with such initialization, which is caused by the lack of abundant observed transitions in experience replay. The start position in experimental map is surrounded with obstacles so that the robot is totally unable to escape from such dilemma. Therefore, rich history experience, including success and failure, is needed to carry abundant information for robot to learn the strategy.

\begin{figure}[htbp]
\centerline{\includegraphics[width=0.37\textwidth]{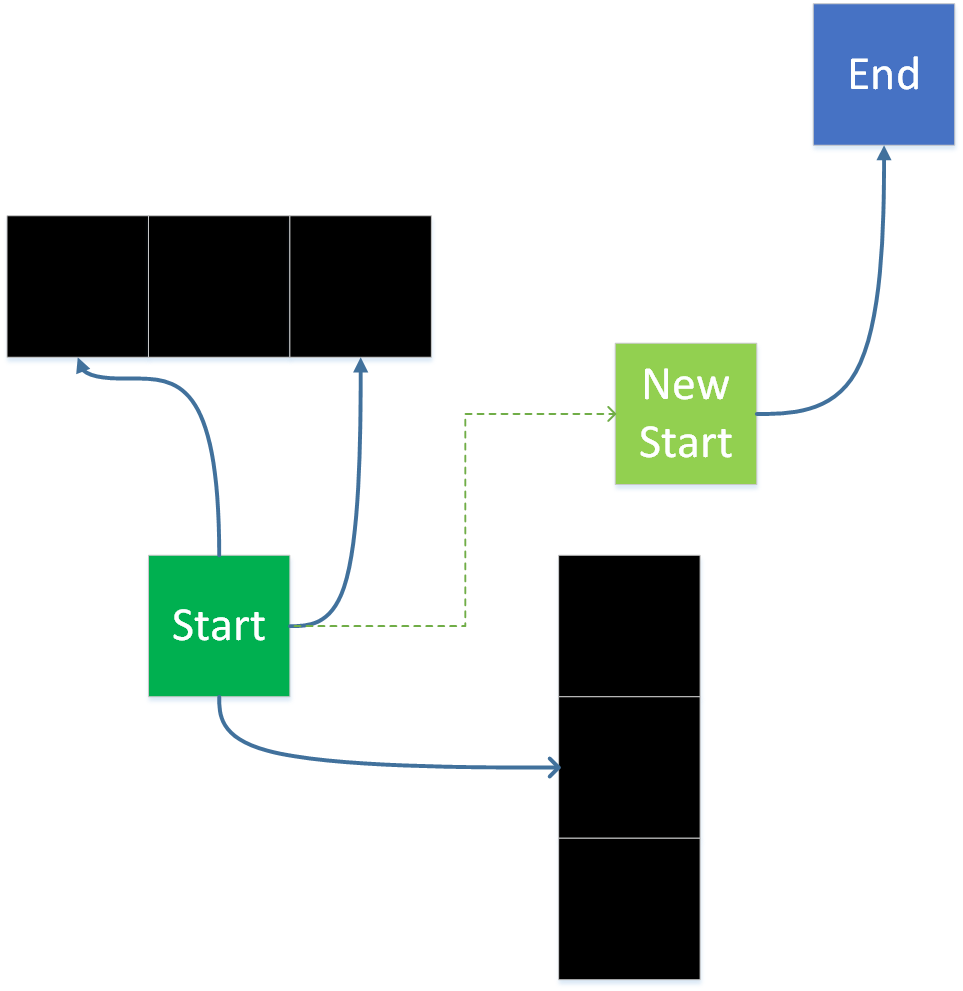}}
\caption{The initialization of the start point during the training.}\label{Fig. 4}
\end{figure}

In Rapidly-Exploring Random Tree (RRT) \cite{b3} a tree to find an efficient path is formed from the start position to the end position. The tree rooted in start point grows via using random samples from the map. If the connection between the sample and the nearest state in the tree is feasible, then the sample is the new state of the tree. As showed in Fig. 4, robot could find a way to escape from the dilemma if it can find the proper point. However, robot could presumably be trapped in such dilemma since the proper point is unknown before the training. To help robot obtain such state without man-made interference, a random value $p_r$ ranged from $[0,1]$ serves to decide robot to initialize to whether the start position or one of the free positions (the new start in Fig. 4) after each ending. In this way, many efficient action-state transitions might be easily formed to achieve the abundance in experience replay.

\subsection{Designed Reward Function}
Another reason that robot could not learn the strategy might be that the reward is identical for all the free positions, resulting in difficult convergence of the network. In order to solve the problem, reward function is specially designed for the application of DQN or Q-learning in path planning in \cite{b21} and \cite{b23}. The main idea of both designed reward functions is to define the reward of the free position as its distance to the end position. The reward of one free position is lower as its distance to the end position is farther. However, such definition to the reward might be not comprehensive because it is more sensible to take the length of path into account. As the definition of the cost in A*  \cite{b1}, the total cost of one free point is the sum of its cost to the start point and to the end point. Therefore, the definition in A* considers the optimal path from the overall situation. In this work, the definition of the reward for the free positions follows as
\begin{equation}
R_t=\alpha(d_{re}(t-1)-d_{re}(t))+\beta(d_{rs}(t)-d_{rs}(t-1)-d_{t,t-1})\label{eq}
\end{equation}

where $\alpha$, $\beta$ are coefficients. $\alpha$ term decides reward by calculating the difference of $d_{re}(t)$, distance between the position at $t$ and the end , and $d_{re}(t-1)$. The reward for the current position ($t$) is higher if its distance to the end position is closer compared to the last position ($t-1$) .

\begin{figure}[htbp]
\centerline{\includegraphics[width=0.4\textwidth]{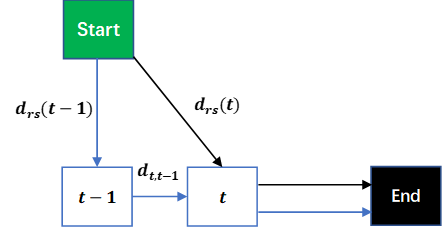}}
\caption{The principle of $\beta$ term in reward function}\label{Fig. 3}
\end{figure}

Fig. 5 shows the principle of $\beta$ term in reward function. $d_{rs}(t)$ refers to the distance between the position at $t$ and the start. There are two paths from the start to the end in Fig. 5 and the blue one is obviously worse than the black one. The cost is zero for black path because the start and the current position is straight-lined connected. Inversely in blue path, the distance between the state at $t$ and the last state at $t-1$ ($d_{t,t-1}$) produces more cost to the path from the start to the current position, which shall negatively influence the reward for the current position.
\section{Experiments}
In this section, the experiments are illustrated respectively according to the three goals mentioned above. The major parameters showed in Table. 1 remain unchanged in the three experiments. Particularly, the distance is measured by Euclidean Distance according to the position of robot. The following improved DDQN approach leverages both the improvements on initialization and reward function since either of them can cause the success in training according to our preliminary tests.
\begin{table}[!htbp]
\caption{Major Parameters in experiments}
\centering
\begin{tabular}{|c|c|c|c|c|c|c|c|}
\hline
Parameters & Mini-batch number & $\gamma$  & Reward for obstacles & Reward for the end & $p_r$ & $\alpha$ & $\beta$\\
\hline
Value & 32 &  0.95 &  -10  & 10 & 0.5 & 0.6 & 0.4\\
\hline
\end{tabular}
\end{table}

\subsection{Planing from the given start to the given end}
For the goal i) and ii), the grid map in Fig. 1 is directly used for the environment of robot. In training process, robot is allowed to constantly interact with the environment and executes the initialization in each round. Fig. 6 shows the training results for the goal i) of the unmodified DDQN and the improved DDQN, which combines  Average reward for each step in each round is used to measure whether the network is successfully trained or not. It is found from Fig. 6(a) that the average reward of the basic DDQN are generally below 0, which means the robot is unable to learn any strategy in such environment. On the contrary, the average reward of the improved DDQN (in Fig. 6(b))are overall increasing until the end of training due to the rich experiments in experience replay. Furthermore, for the improved DDQN, the training step, equally robot's moving step, is 15000 (about 2800 rounds) to achieve the convergence of network. However, robot could not even find a sufficient path with 50000 (about 5600 rounds) training steps in basic DDQN. Thus the improved model significantly reduces the training time.
\begin{figure}[htbp]
\centering    
  \subfigure[]{
    \includegraphics[scale=0.3]{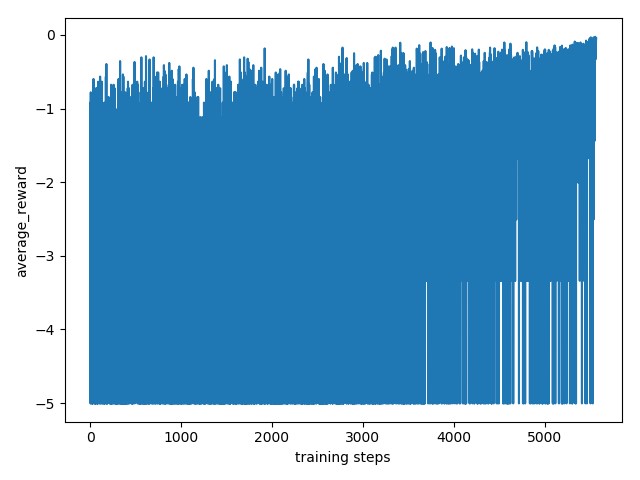}
  }
  \subfigure[]{
    \includegraphics[scale=0.3]{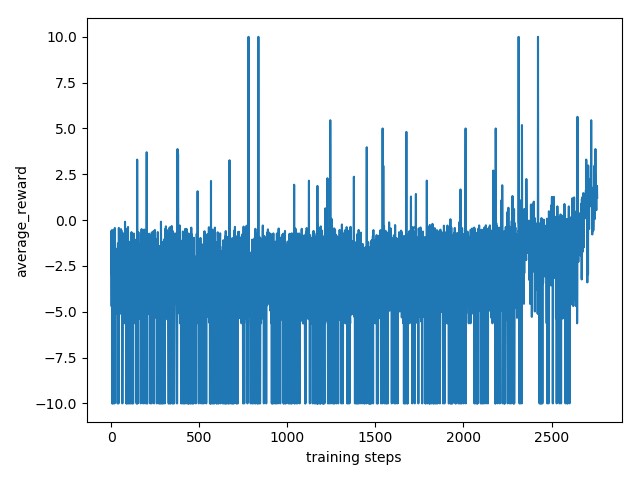}
  }
%
%
\caption{The training results for the goal i) of different algorithms. (a) The training result of DDQN. (b) The training result of the improved DDQN. } 
\label{fig:1}  
\end{figure}

Fig. 7 shows the optimal path planned by the improved DDQN and A*. The path length based on the improved DDQN is 22.97 while the length based on A* is 25.31, which further proves the preponderant performance of the improved DDQN in path planning from the given start to the given end.
\begin{figure}[htb]
  \centering
  \subfigure[]{
    \includegraphics[width=1.6in]{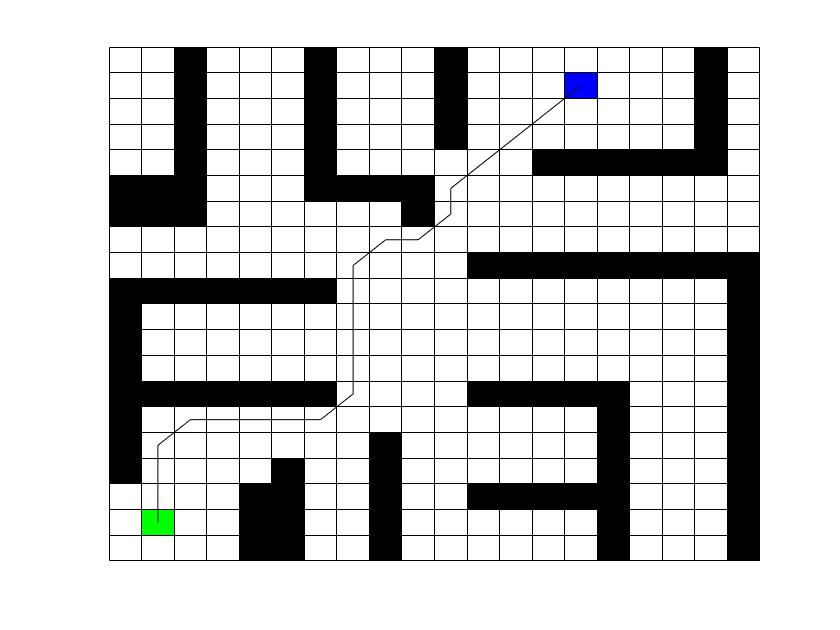}
  }
  \subfigure[]{
    \includegraphics[width=1.6in]{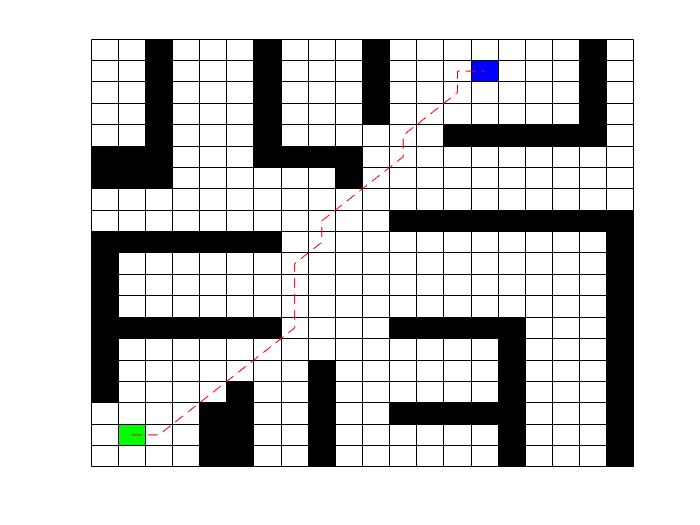}
  }
  \caption{The path planned by two algorithms for goal i). (a) The path planned by A*. (b) The path planned by the improved DDQN.}
\end{figure}
Additionally, the total testing time of the improved DDQN is about 0.15 seconds, which can meet the need for some online projects.

For the goal ii), the trained model could be directly tested. In experiment map, the total number of the valid free positions is 286 (10 free positions are trapped in dilemma) and all of them are tested under the trained model. According to the result, it is found that robot is familiar with the current map since it can plan a valid or optimal path from 199 free positions (nearly 70\%) as the given start to the given end. Fig. 8 shows three testing results for the goal ii) and robot could path an efficient path from different start points located in the dilemma.

\begin{figure}[htb]
  \centering
  \subfigure[]{
    \includegraphics[width=1.4in]{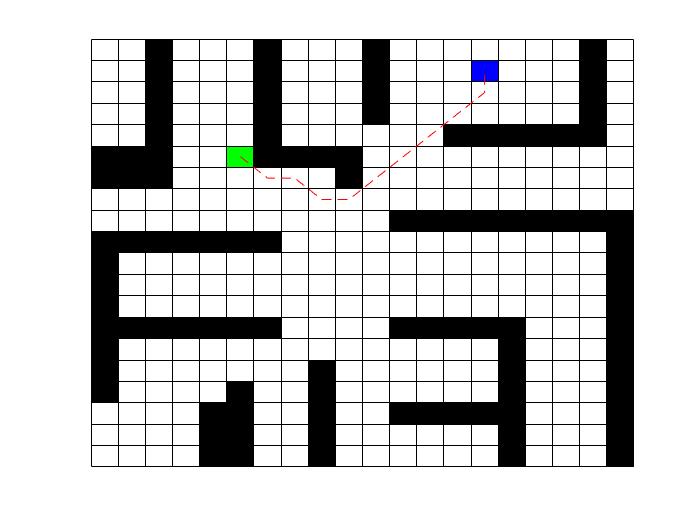}
  }
  \subfigure[]{
    \includegraphics[width=1.4in]{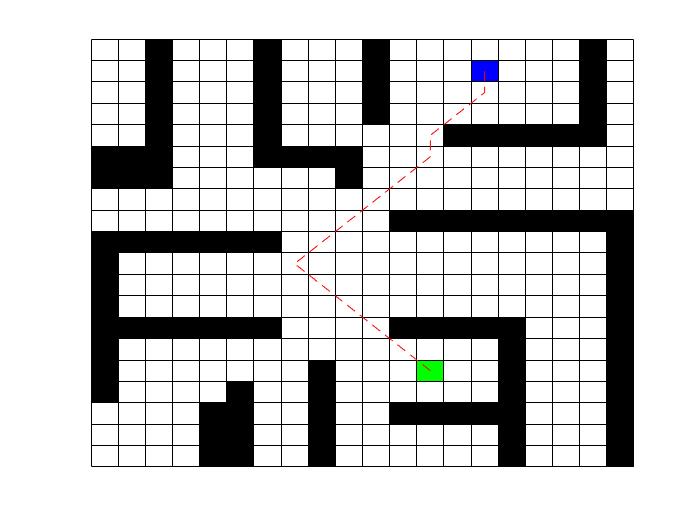}
  }
    \subfigure[]{
    \includegraphics[width=1.4in]{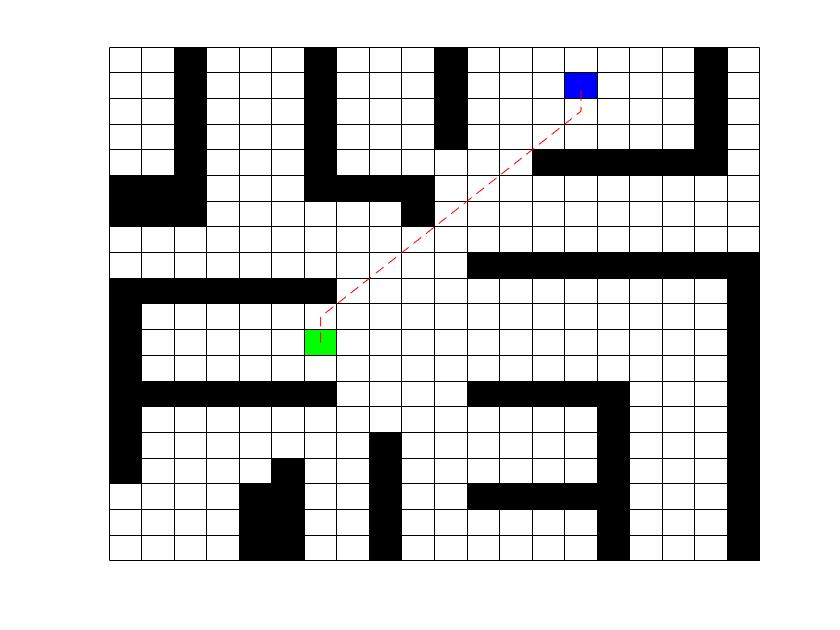}
  }
  \caption{The paths planned from the different given starts to the given end.}
\end{figure}
\subsection{Planing with static noises}
In this experiment, new static obstacles are particularly set on part of free positions that make up the optimal path in Fig. 7(b), which could be treated as the disturbance or noises. The number of new obstacles is a random integer in interval $[1,5]$ and 300 maps are generated by randomly setting new obstacles to the required free points near the optimal trajectory based on the goal i).To train the network, 100 maps are chosen to be the training samples and the other 200 maps are testing samples. According to the trained model with the training steps to 120000, robot can successfully plan an efficient path in 158 testing samples (nearly 80\%) and in 98 training samples.

\begin{figure*}[htbp]
  \centering
  \subfigure[]{
    \includegraphics[width=1.4in]{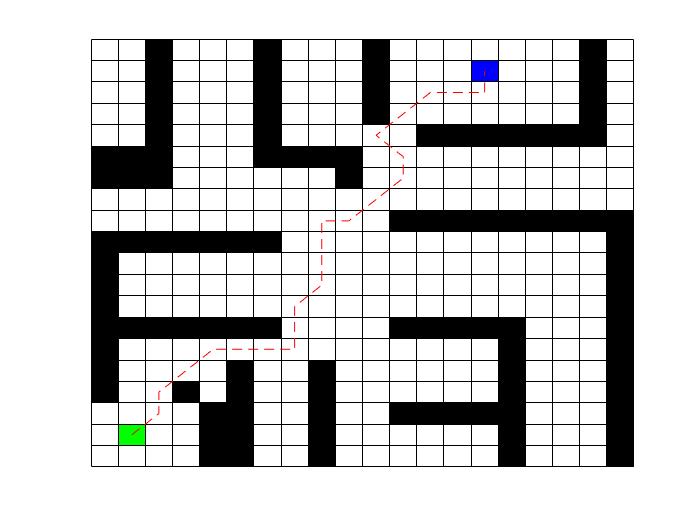}
  }
  \subfigure[]{
    \includegraphics[width=1.4in]{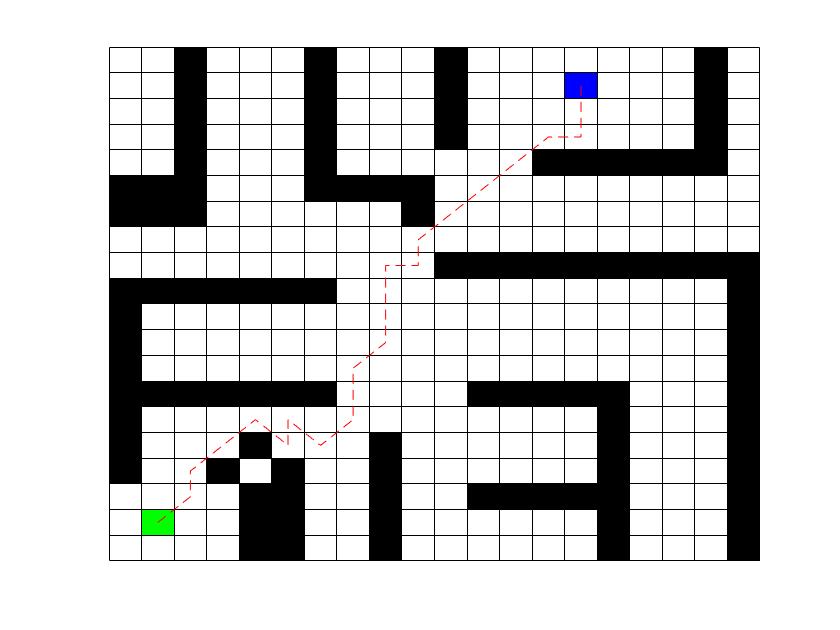}
  }
    \subfigure[]{
    \includegraphics[width=1.4in]{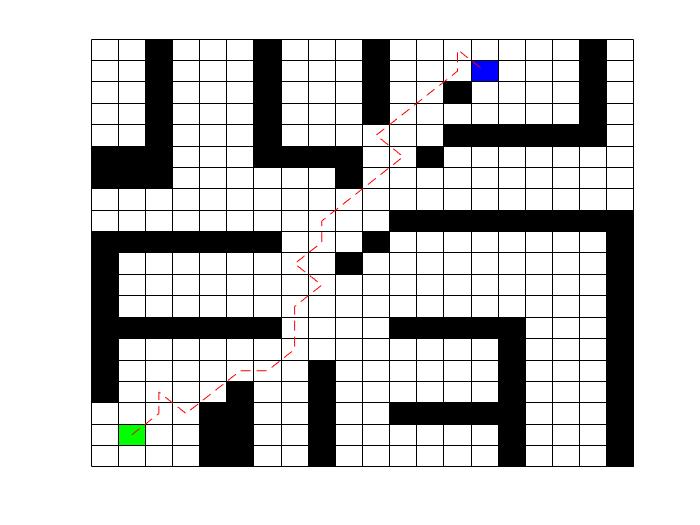}
  }
      \subfigure[]{
    \includegraphics[width=1.4in]{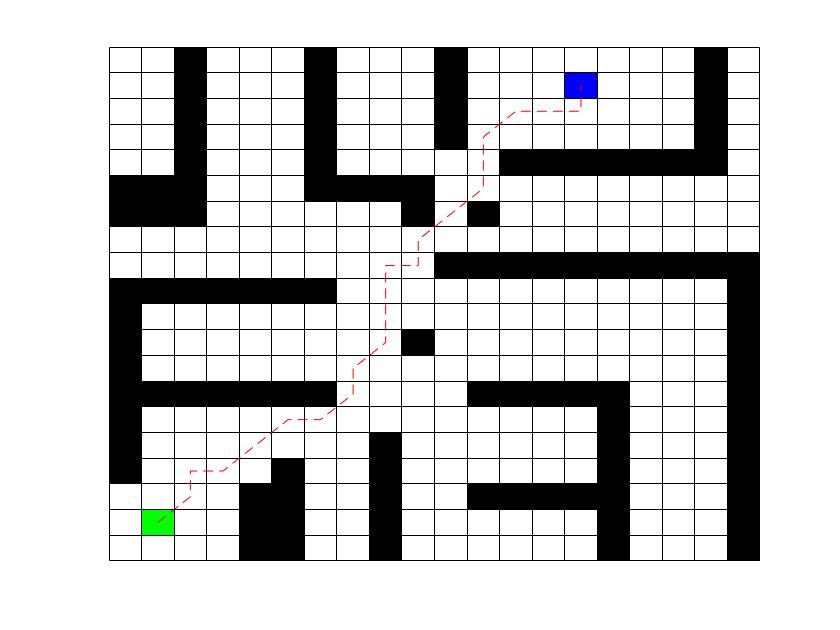}
  }
      \subfigure[]{
    \includegraphics[width=1.4in]{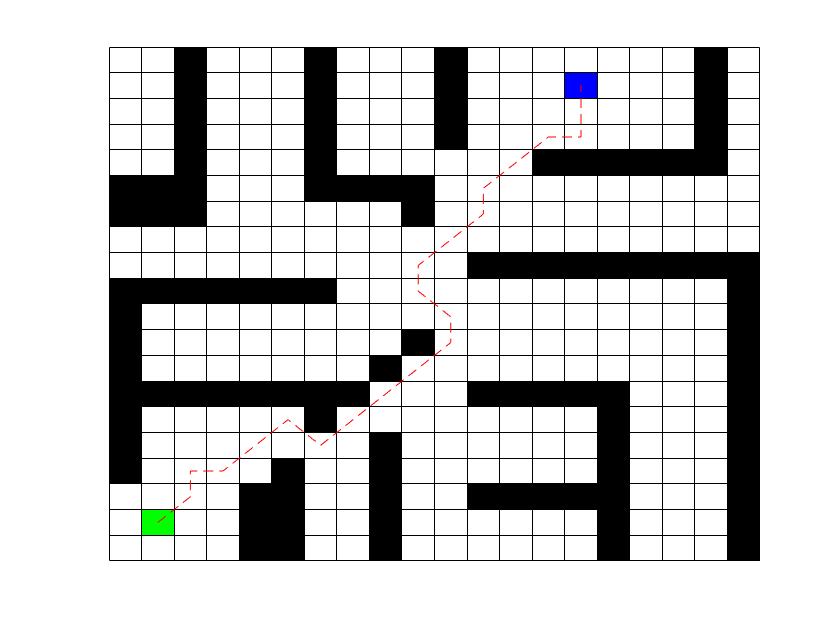}
  }
      \subfigure[]{
    \includegraphics[width=1.4in]{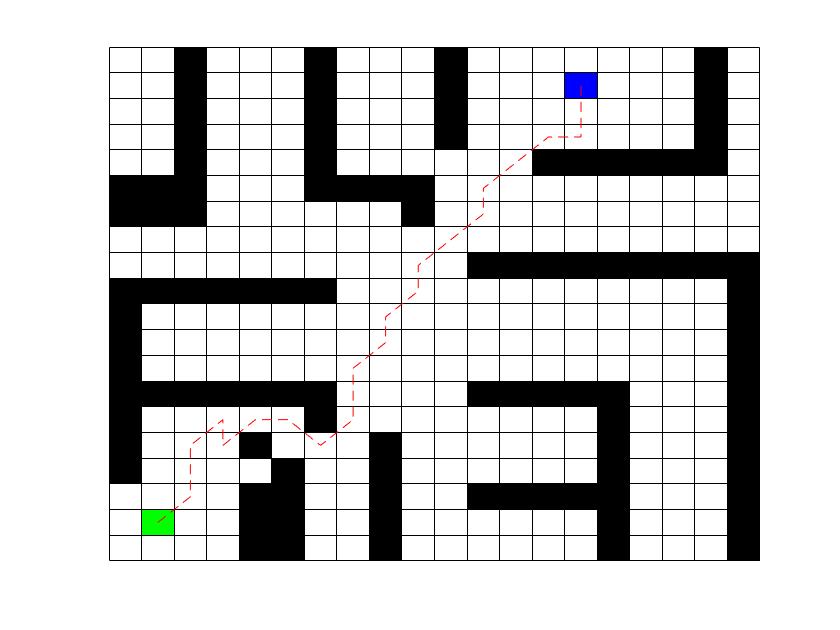}
  }
  \caption{The paths planned from the given start to the given end with noises.}
\end{figure*}
In Fig. 9(a) and (b), disturbing points are mainly set around the give start and hinder the robot optimal moving direction. But robot can bypass such obstacle forehead and move forward to the end. Meanwhile, in Fig. 9(c) and (d), the new obstacles around the given end are also ignored by the robot. Even in Fig.9(e), the noises consist of a "ladder" form originated from the dilemma around the start. Surprisingly but reasonably, the robot can avoid the noises and successfully find an efficient path towards the target. In Fig. 9(f), the robot could also escape the hard dilemma. As model is simply trained with 100 samples, the robot could succeed in nearly 80\% testing samples. According to the results, the rationality of improvements on the initialization and reward function can be directly attested and based on which robot can robustly plan a path from the given start to the given end in an environment.
\section{Conclusion}
The problem particularly addressed is that the robot cannot plan an efficient path from a given start to a given end in complex environment based on DDQN. This paper presents an improved DDQN algorithm to be specifically applied in robot path planning by learning theories from A* and RRT. For the improvements on DDQN, the initialization of the robot during the training is redefined so as to solve the problem of lacking experiments. Importantly, such change emphasizes to make robot learn the efficient experiments without deliberate generation of good experience. In addition, the reward function for the free position is specially designed by reference to A* in order to accelerate the convergence of network. Three goals are introduced in the experiments for the robot path planing, and the simulation results show that robot could learn the strategy to safely and robustly reach the end position based on the improved DDQN.
%
%

\end{document}